# Load and Renewable Energy Forecasting Using Deep Learning for Grid Stability

Dr. Kamal Sarkar
Professor
CSE Dept.,
Jadavpur University
jukamal2001@yahoo.com

**Abstract**

*As the energy landscape changes quickly, grid operators face several challenges, especially when integrating renewable energy sources with the grid. The most important challenge is to balance supply and demand because the solar and wind energy are highly unpredictable. When dealing with such uncertainty, trustworthy short-term load and renewable energy forecasting can help stabilize the grid, maximize energy storage, and guarantee the effective use of renewable resources.*

*Physical models and statistical techniques were the previous approaches employed for this kind of forecasting tasks. In forecasting renewable energy, machine learning and deep learning techniques have recently demonstrated encouraging results. More specifically, the deep learning techniques like CNN and LSTM and the conventional machine learning techniques like regression that are mostly utilized for load and renewable energy forecasting tasks. In this article, we will focus mainly on CNN and LSTM-based forecasting methods.*

## 1. Introduction

Fossil fuels have been the world's main sources of electric energy. Hydrocarbons or their derivatives, including petroleum, natural gas, and coal, are referred to as fossil fuels. Fossil fuels take millions of years to develop, and the known viable supplies are running out far more quickly than they are being created. Fossil fuels additionally emit greenhouse gases into the atmosphere, expediting climate change and harming the environment that supports human life. As a result, for the development of more sustainable energy systems, renewable energy has received much attention in recent years Renewable energy such as solar energy, wind power, tidal energy, and geothermal energy, can be can be recycled in nature.

Electricity in the globe has mostly come from fossil fuels. Fossil fuels are hydrocarbons or their derivatives, such as coal, natural gas, and petroleum. The development of fossil fuels takes millions of years , and the known feasible supplies of fossil fuels are running out significantly more quickly than they are being created,. In addition, greenhouse gases released into the atmosphere by fossil fuels accelerate climate change and damage the ecosystem that sustains human life. As a result, renewable energy has drawn a lot of attention recently for the creation of more sustainable energy systems. Renewable energy sources that can be recycled in nature include solar, wind, tidal, and geothermal energy.

Compared to fossil fuels, renewable energy offers a number of significant advantages. First, its resources are limitless and renewable; second, it is clean, green, and low carbon, making it less harmful to the environment. The risk of atmospheric pollution is also decreased by lowering sulfur, carbide, and dust, and (3) appropriate use of renewable energy can lessen the exploitation of natural resources. Many nations, like China and the United States, have established laws to promote support renewable energy because of its immediate advantages [1].

Although renewable energy is considered the most viable alternative to fossil fuels, especially when integrated on a broad scale, the uncertainty surrounding it has a substantial impact on the stability and reliability of energy systems. Moreover, renewable energy's high variability, irregularity, and unpredictability increase electric energy networks' reserve capacity—which is the difference between plant capacity and maximum demand—and increase the cost of producing electricity. Consequently, forecasting for renewable energy is crucial for reducing the uncertainty associated with renewable energy sources [2]. For instance, wind energy management and grid integration benefit greatly from wind power forecasts.

Several algorithms that reliably forecast renewable energy for the next few minutes to the next few days has already been published in the research papers. Long-term, medium-term, and short-term predictions are the three primary types of load forecasting techniques that are crucial for modern grid reliability[3].

Physical approaches, statistical methods, machine learning and deep learning methods, and hybrid methods are the four groups into which these algorithms can be categorized [3]. Physical approaches, which were mostly used for weather prediction (NWP), model atmospheric dynamics based on physical principles and boundary conditions [4]. This type of system used geographical and meteorological data, such as orography, temperature, pressure, and jaggedness, as input.

In early research works, statistical models were applied to uncover underlying hidden statistical patterns in online time series data of renewable energy [5]. The following popular statistical techniques have been published in the literature: the Markov Chain model [9], the Kalman filter [8], the auto regressive moving average [7], and the Bayesian approach [6]. Although these methods showed encouraging results, they were not free from flaws. Since forecasting data is time series data, their primary flaw is their limitations in processing time series.

Due to the potential capabilities of AI algorithms for identifying important patterns from high dimensional data and learning intricate nonlinear relationships between input features and output, the development of Artificial Intelligence (AI) techniques, particularly machine learning and deep learning-based forecasting models, has demonstrated superior performance over physical methods and statistical approaches [10]. Additionally, they are capable of handling a variety of data kinds, including time series, geographic, and meteorological data.

Large power systems' stability and dependability rely on precise forecasts of both supply and future demand [11]. By precisely forecasting load demand and renewable energy generation, long short-term memory (LSTM) neural networks can assist in resolving difficulties of integration of renewable energy sources with the power grid [12]. Better planning and control of power generation resources are made possible by these forecasts, guaranteeing an ideal supply and demand balance. Because LSTM networks can identify long-term trends and dependencies in time series data, they are especially well-suited to managing the erratic and intermittent character of renewable energy sources like wind and solar power [13]. Grid operators can make well-informed judgments about dispatching and adjusting conventional generators by using LSTMs to forecast load demand and generation from renewable sources, guaranteeing frequency stabilization and system dependability.

## 2. Deep Learning Methods for Load and Renewable Energy Forecasting

A class of machine learning models known as artificial neural networks (ANNs) draws inspiration from the biological neural networks found in the human body [14]. An input layer, one or more hidden layers, and an output layer are the three essential parts of an artificial neural network. Both classification and regression problems can be solved with ANN. Artificial Neural Networks (ANNs) can be classified as shallow or deep neural networks based on the number of hidden layers. Deep neural networks are ANNs with more than two hidden layers. An ANN presents a complex function that converts the input vector that is received at the input layer into either a predicted value for regression or an output vector for classification. An ANN can map non-linearly the input to the output.

Every node in a layer of a fully connected feedforward neural network is connected to every other node in the next layer. The relationship between a node at layer L and another node at layer L+1 is indicated by a weight parameter. These connection weights are an ANN model's parameters. The training set, which consists of a collection of (x, y) pairs, where x is the input feature vector and y is the intended output value(s), is used to learn the weight parameters using the backpropagation (BP) algorithm [15] in supervised mode. During training, the BP algorithm adjusts the network's weights to make sure the network produces the desired result for a particular training input. In order to minimize the overall loss (error) on the training set, the BP algorithm uses the gradient of the loss function, (which is a function of the weights) to find the best possible values of the weight parameters at which error becomes minimum.

When making a prediction, an unseen input instance is fed into the trained network. Each node's values are calculated, and the output node provides the predicted y value. Figure 2 depicts a small shallow neural network. Non-linear activation functions, such as Sigmoid, are used by the output nodes and all hidden nodes to guarantee non-linearity. ReLU (rectified linear unit) activation functions are used to the hidden nodes for a very data-intensive task in order to speed up training and prevent overfitting.

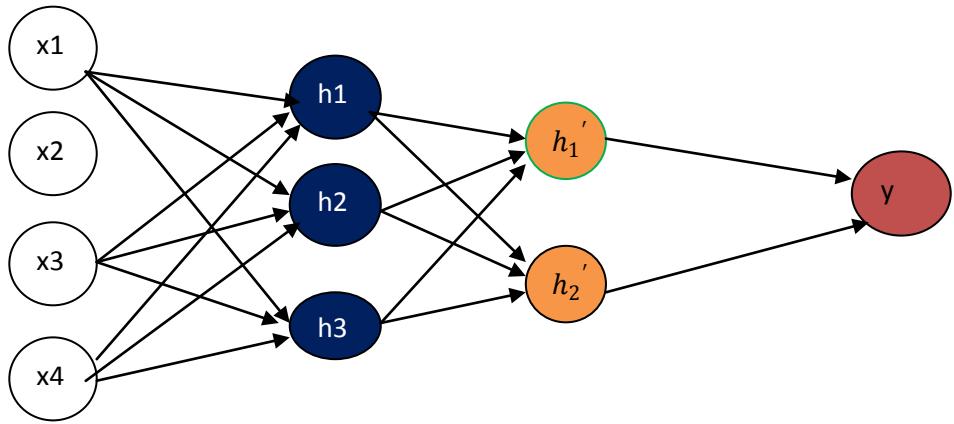

**Figure 2. A small shallow neural network with two hidden layers**

Some researchers have already investigated the use of artificial neural networks in forecasting renewable energy. Global solar radiation has been predicted using ANNs [16][17][18]. Additionally, ANN was used to anticipate load demand and wind power generation. An ANN was used in the study presented in [19] to forecast wind power generation and load demand. In this work, the various meteorological parameters—such as temperature, wind speed, and atmospheric pressure—were utilized as input features. An artificial neural network (ANN) model was presented by Chen et al. (2019) [20] for the short-term prediction of wind power. They took into account temperature, relative humidity, atmospheric pressure, wind direction, and wind speed at different elevations.

Although shallow artificial neural networks (ANNs) were used for forecasting renewable energy, deep learning has garnered a lot of interest recently because of its great generalization skills, automatic feature learning, and huge data handling capabilities. The term "deep" describes a deep neural network's multiple layers and the several intermediate steps used by it in computing path from input to output. A most effective and popular method for many applications, including computer vision, speech recognition, machine translation, etc., is deep learning. Deep neural networks come in a variety of forms nowadays. Among them, deep belief networks (DBNs), convolutional neural networks (CNNs), long short-term memory (LSTM) networks, autoencoders, and combinations of these deep networks have been used for renewable energy forecasting.

LSTM is a deep neural network which is suitable for learning sequence data. It is an improved version of RNN (Recurrent Neural Networks). With fewer parameters than LSTM, GRU (Gated Neural Network) is quite similar to LSTM. Therefore, compared to LSTM, GRU has a quicker training time and a less severe overfitting issue. For wind power prediction, Kisvari et al. (2021)[21] suggested a GRU-based approach.

Using an adjustable learning rate, an RNN-based model was used in [22] for forecasting daily, mean monthly, and hourly sun radiation. In this work, authors used meteorological data to construct its models. The RNN performed better than the shallow feedforward neural network, according to the authors' findings.

A hybrid deep learning model that integrates CNN and LSTM for reliable power generation forecasting in photovoltaic (PV) systems was proposed by Lim et al. (2022)[23]. They gathered data on PV power generation from a plant in Busan, Korea, and took into account a number of environmental parameters, including temperature and sun radiation. While LSTM considers the long temporal dependency, CNN learns local features from the input sequence. As a result, the CNN-LSTM model may outperform the CNN or LSTM-based models alone. Additionally, Wu et al. (2020)[24] suggested a CNN-LSTM model for very short-term wind power forecasting. For efficient wind power forecasting in a very short amount of time, this model uses CNN to extract spatial correlation features and LSTM to extract temporal correlation features from the input data.

An unsupervised deep learning model called Autoencoder can be used to extract discriminative and effective features from a lot of unlabeled input. It is frequently applied to data dimension reduction and feature extraction. Autoencoder has been used by some researchers to forecast renewable energy. In this case, autoencoder was used to extract significant features from the input data such as weather information, historical energy production statistics, and other pertinent data. For feature extraction, Dairi et al. (2015)[25] employed a variational autoencoder (VAE) model. Using layer-wise training, the predictive model was first trained in unsupervised mode. It was then refined using training sub-sequences generated from time series data, where each sub-sequence has the format <x(t), x(t+1)>, where x(t) is the subsequence at time t and y value equals x(t+1), which is the output at time step t that is one step ahead.

For wind speed forecasting, Jaseena and Kovoor (2015)[26] use a hybrid strategy that combines LSTM and an autoencoder. To extract significant features from the input data, the model makes use of an autoencoder. The retrieved features are then used to train the LSTM model for wind speed prediction.

The methods discussed above were either used for load prediction or renewable energy forecasting. But very recently, Saxena et.al. (2024) proposed a bidirectional LSTM-based model for forecasting both load and renewable energy. They advocated that short-term load and renewable energy forecasting both are required to help stabilize the grid, maximize energy storage, and guarantee the effective use of renewable resources.

## 3. Conclusion

Renewable energy sources such as wind and solar power are appealing alternatives that can be used to help fulfil global energy demands. However, because renewable energy sources are unpredictable and variable, reliable forecasting of renewable energy remains a difficult task. Because of this, energy system operators encounter challenges in maintaining the grid's stability and dependability.

Forecasting models for load and renewable energy based on machine learning and deep learning can help grid operators manage the system efficiently. Although there is still room for development, this survey shows that the most advanced ML and DL-based energy forecasting models available today are efficient. SVMs and XGBoost models are shown to be the most effective machine learning algorithms for forecasting renewable energy. CNN-LSTM deep learning models have also demonstrated efficacy. It is also reported that Bi-LSTM model for forecasting both load and renewable energy is effective.

Forecasting renewable energy still faces some obstacles, even with advancements in machine learning and deep learning algorithms. The absence of high-quality training and validation data is one of the major obstacles. Publicly accessible data for projecting renewable energy is frequently noisy and lacking. Since safety and security are important concerns in electric power systems, the data used to build the models should be real. Data must be standardized, and rules and procedures for data exchange must be established in order to support research. This is crucial since developing precise, trustworthy, and secure energy forecasting models is challenging due to the lack of real data.

The goal of future research may be to encourage business and academic cooperation in order to create efficient machine learning and deep learning models and implement them in real-time renewable energy forecasts. Forecasting for renewable energy can be improved by combining various data types gathered from many sources, such as weather, grid, and other external data, through data fusion, and using it in ML and DL models.